\pdfoutput=1

\documentclass[11pt]{article}
\usepackage{acl}

\usepackage{times}
\usepackage{latexsym}

\usepackage[T1]{fontenc}

\usepackage[utf8]{inputenc}

\usepackage{microtype}

\usepackage{url}

\usepackage{booktabs}
\usepackage{graphicx}
\usepackage{multirow}
\usepackage{microtype}
\usepackage{footnote}
\usepackage{amsmath}
\usepackage{amsfonts}
\usepackage{amssymb}
\usepackage{enumitem}
\usepackage{subcaption}
\usepackage{bm}

\makesavenoteenv{tabular}
\makesavenoteenv{table}
\makeatletter
  \newcommand\smallernormal{\@setfontsize\smallernormal{9.6pt}{9.6}}
\makeatother
\makeatletter
  \newcommand\smallernormalAppendix{\@setfontsize\smallernormalAppendix{11pt}{10}}
\makeatother

\title{Natural Language Inference with Self-Attention for\\ Veracity Assessment of Pandemic Claims}

 \author{Miguel Arana-Catania\textsuperscript{1,2}, Elena Kochkina\textsuperscript{3,2}, Arkaitz Zubiaga\textsuperscript{3}, Maria Liakata\textsuperscript{3,2}, \\ 
 \textbf{Rob Procter\textsuperscript{1,2}, Yulan He\textsuperscript{1,2}} \\
         \textsuperscript{1}Department of Computer Science, University of Warwick, UK \\ \textsuperscript{2}The Alan Turing Institute, UK \\ \textsuperscript{3}Queen-Mary University of London, UK}
         
\date{}

\begin{document}
\maketitle
\begin{abstract}
We present a comprehensive work on automated veracity assessment from dataset creation to developing novel methods based on Natural Language Inference (NLI), focusing on misinformation related to the COVID-19 pandemic. We first describe the construction of the novel PANACEA dataset consisting of heterogeneous claims on COVID-19 and their respective information sources. The dataset construction includes work on retrieval techniques and similarity measurements to ensure a unique set of claims. We then propose novel techniques for automated veracity assessment based on Natural Language Inference including graph convolutional networks and attention based approaches. We have carried out experiments on evidence retrieval and veracity assessment on the dataset using the proposed techniques and found them competitive with SOTA methods, and provided a detailed discussion.
\end{abstract}

\section{Introduction}

In recent years, and particularly with the emergence of the COVID-19 pandemic, significant efforts have been made to detect misinformation online with the aim of mitigating its impact. With this objective, researchers have proposed numerous approaches and released datasets that can help with the advancement of research in this direction.

Most existing datasets \cite{d2021fake} focus on a single medium (e.g., Twitter, Facebook, or specific websites), a unique information domain (e.g., health information, general news, or scholarly papers), a type of information (e.g., general claims or news), or a specific application (e.g., verifying claims, or retrieving useful information). This inevitably results in a limited focus on what is a complex, multi-faceted phenomenon. With the aim of furthering research in this direction, the contributions of our work are twofold: (1) creating a new comprehensive dataset of misinformation claims, and (2) introducing two novel approaches to veracity assessment.

In the first part of our work, we contribute to the global effort on addressing misinformation in the context of COVID-19 by creating a dataset for PANdemic Ai Claim vEracity Assessment, called the \textbf{PANACEA dataset}. It is a new dataset that combines different data sources with different foci, thus enabling a comprehensive approach that combines different media, domains and information types. 
To this effect our dataset brings together a heterogeneous set of \emph{True} and \emph{False} COVID claims and online sources of information for each claim.  The collected claims have been obtained from online fact-checking sources, existing datasets and research challenges. We have identified a large overlap of claims between different sources and even within each source or dataset. Thus, given the challenges of aggregating multiple data sources, much of our efforts in dataset construction has focused on eliminating repeated claims. Distinguishing between different formulations of the same claim and nuanced variations that include additional information is a challenging task. Our dataset is presented in a large and a small version, accounting for different degrees of such similarity. Finally, the homogenisation of datasets and information media has presented an additional challenge, since fact-checkers use different criteria for labelling the claims, requiring a specific review of the different kinds of labels in order to combine them.

In the second part of our work, we propose NLI-SAN and NLI-graph, two novel veracity assessment approaches for automated fact-checking of the claims. Our proposed approaches are centred around the use of Natural Language Inference (NLI) and contextualised representations of the claims and evidence. NLI-SAN combines the inference relation between claims and evidence with attention techniques, while NLI-graph builds on graphs considering the relationship between all the different pieces of evidence and the claim.

Specifically we make the following contributions:

\begin{itemize}
 \itemsep0em
 \item We describe the development of a comprehensive COVID fact-checking dataset, PANACEA, as a result of aggregating and de-duplicating a set of heterogeneous data sources. The dataset is available in the project website\footnote{\url{https://panacea2020.github.io/} \url{https://doi.org/10.5281/zenodo.6493847}}, as well as a fully operational search platform to find and verify COVID-19 claims implementing the proposed approaches.
 \item We propose two novel approaches to claim verification, NLI-SAN and NLI-graph.
 \item We perform an evaluation of both evidence retrieval and the application of our proposed veracity assessment methods on our constructed dataset. Our experiments show that NLI-SAN and NLI-graph have state-of-the-art performance on our dataset, beating GEAR \citep{zhou2019gear} and matching KGAT \citep{liu2019fine}. We discuss challenging cases and provide ideas for future research directions.
\end{itemize}

\section{Related Work}
\paragraph{COVID-19 and misinformation datasets.}  Comprehensive information on COVID-19 datasets is provided in Appendix \ref{appendix_sources}. Such datasets include the CoronaVirusFacts/DatosCoronaVirus Alliance Database, the largest existing collection of COVID claims and the largest existing network of journalists working together on COVID misinformation, an essential reference for our work; COVID-19-TweetIDs \citep{chen2020tracking} the widest dataset of COVID tweets with more than 1 billion tweets; Cord-19: The COVID-19 open research dataset \citep{wang2020cord}, the largest downloadable set of scholarly articles on the pandemic with nearly 200,000 articles. 
General misinformation datasets linked to our verification work include: Emergent \citep{ferreira2016emergent} collection of 300 labeled claims by journalists; LIAR \citep{wang2017liar} with 12,836 statements from PolitiFact with detailed justifications; FakeNewsNet \citep{shu2020fakenewsnet} collecting not only claims from news content, but also social context and spatio-temporal information; NELA-GT-2018 \citep{norregaard2019nela} with 713,534 articles from 194 news outlets; FakeHealth \citep{dai2020ginger} collecting information from HealthNewsReview, a project critically analysing claims about health care interventions; PUBHEALTH \citep{kotonya2020explainable} with 11,832 claims related to health topics; FEVER \citep{Thorne18Fever} as well as its later versions FEVER 2.0 \citep{Thorne19FEVER2} and FEVEROUS \citep{Aly21Feverous}, containing claims based on Wikipedia and therefore constituting a well-defined, informative and non-duplicated information corpus; SciFact \citep{wadden2020fact} also from a very different domain, containing 1,409 scientific claims. Our dataset is a real-world dataset bringing together heterogeneous sources, domains and information types.

\paragraph{Approaches to claim veracity assessment.} We employ our dataset for automated fact-checking and veracity assessment \cite{zeng2021automated}. Researchers such as \citet{hanselowski2018ukp, yoneda2018ucl, luken2018qed, soleimani2020bert, pradeep2021vera} analysed the veracity relation between the claim and each piece of evidence independently, combining this information later. Other authors considered multiple pieces of evidence together \citep{Thorne18Fever,nie2019combining,stammbach2019team}. Different pieces of evidence have been previously combined using graph neural networks \citep{zhou2019gear, liu2019fine,zhong2019reasoning}. Many of these authors have centred their techniques on the use of NLI  \citep{chen2016enhanced,ghaeini2018dr,parikh2016decomposable,li2019several} to verify the claim. In our work we also make use of NLI results of claim-evidence pairs, but propose alternative approaches built on a self-attention network and a graph convolutional network for veracity assessment.

\section{Dataset Construction}
This section describes our dataset construction by selecting COVID-19 related data sources (\textsection{\ref{sec:dataSource}}), 
and applying information retrieval and re-ranking techniques to remove duplicate claims (\textsection{\ref{sec:preprocessing}}). 

\begin{table*}[!ht]\small
    \centering
    \begin{tabular}{p{3cm}p{7cm}p{2cm}p{2cm}}
    \toprule \textbf{Data Source} & \textbf{Description} & \textbf{Domain} & \textbf{No. of claims} \\
    & & & \textbf{(False / True)}\\
    \midrule
CoronaVirusFacts Database &  Published by Poynter, this online source combines fact-checking articles from more than 100 fact-checkers from all over the world, being the largest journalist fact-checking collaboration on the topic worldwide. & Heterogeneous & 11,647 \newline (11,647 / 0) \\
\midrule
CoAID dataset \newline \citep{cui2020coaid} & This contains fake news from fact-checking websites and real news from health information websites, health clinics, and public institutions. & News &  5,485 \newline (953 / 4,532) \\
\midrule
MM-COVID \newline\citep{li2020mmcovid} & This multilingual dataset contains fake and true news collected from Poynter and Snopes. & News & 3,409 \newline (2,035 / 1,374) \\
\midrule
CovidLies \newline\citep{hossain2020covidlies} & This contains a curated list of common misconceptions about COVID appearing in social media, carefully reviewed to contain very relevant and unique claims. & Social media & 62 \newline (62 / 0) \\
\midrule
TREC Health Misinformation track & Research challenge using claims on the health domain focused on information retrieval from general websites through the Common Crawl corpus (commoncrawl.org). & General \newline websites & 46 \newline (39 / 7) \\
\midrule
TREC COVID challenge \newline\cite{voorhees2021trec,roberts2020trec} & Research challenge using claims on the health domain focused on information retrieval from scholar peer-reviewed journals through the CORD19 dataset \citep{wang2020cord}, the largest existing compilation of COVID-related articles. & Scholar papers &  40 \newline (3 / 37) \\
    \bottomrule
    \end{tabular}
    \caption{Data sources used for the construction of our dataset. The last column shows the number of claims before de-duplication.}
    \label{tab:dataSources}
\end{table*}

\subsection{Data Sources}
\label{sec:dataSource}

We first identified a set of COVID-19 related data sources to build our dataset. Our aim is to have the largest compilation of non-overlapping, labelled and verified claims from different media and information domains (Twitter, Facebook, general websites, academia), and used for different applications (media reporting, veracity evaluation, information retrieval challenges, etc.). We have included any large dataset or media, to our knowledge, related to that objective that includes claims together with their information sources. The data sources identified are shown in Table~\ref{tab:dataSources}. More details and pre-processing steps are presented in Appendix \ref{appendix_sources}. By processing and combining these sources we obtained 20,689 initial claims.

\subsection{Claim De-duplication}
\label{sec:preprocessing}
We processed claims and removed: exact duplicates; claims making only a direct reference to existing content in other media (audio, video, photos); automatically obtained content not representing claims; entries with claims or fact-checking sources in languages other than English.

The similarity of claims was then analysed using: BM25 \cite{robertson1995okapi,crestani1998document,robertson2009probabilistic} and BM25 with MonoT5 re-ranking \cite{nogueira2020document}. BM25 is a commonly-used ranking function that estimates the relevance of documents to a given query. MonoT5 uses a T5 model trained using as input the template `\texttt{Query:[query] Document:[doc] Relevant:}', fine-tuned to produce as output the token `\emph{True}' or `\emph{False}'. A softmax layer applied to those tokens gives the respective relevance probabilities. These methods are used to identify not only claims similar in content, but also distinct claims that are sufficiently relevant when searching for information about them. This ensures that the claims presented are unique, and avoids overlap between training and testing cases when using the data to train veracity assessment models. These methods were carried out using Pyserini\footnote{\url{https://github.com/castorini/pyserini}} and PyGaggle\footnote{\url{https://github.com/castorini/pygaggle}}. The set of claims was indexed and a search was performed for each of the claims to detect similar claims. We created two versions of the dataset by varying the similarity threshold between claims. 
The \textsc{Large} dataset excludes claims with a 90\% probability of being similar, while in the \textsc{Small} dataset the probability is increased to 99\%, as obtained through the MonoT5 model. These thresholds were chosen empirically by manual inspection of the results with simultaneous consideration of the efficiency of the method.

As a further assessment of the uniqueness of the claims, we evaluated the de-duplication process using BERTScore\footnote{ \url{https://github.com/Tiiiger/bert_score}} \citep{zhang2019bertscore} on the resulting datasets. We used the linked code with a RoBERTa-large model with baseline rescaling. We compared each claim with all the other claims in the dataset and kept the score of the most similar match. The mean and standard deviation, and the 90th percentile of claim similarity values are shown in the upper part of Table \ref{tab:P-dataset}. The average claim similarity has been drastically reduced in the \textsc{Large} dataset compared to the original dataset and further reduced in the \textsc{Small} dataset.

\begin{table}[!ht]\small
    \centering
    \begin{tabular}{p{7.3cm}}
    \toprule \textbf{Claim 1:} Losing your sense of smell may be an early symptom of COVID-19. \\
\midrule
\textbf{Exclude from \textsc{Large} and \textsc{Small}}:\\
Loss of smell may suggest milder COVID-19. \\
\textbf{Exclude from \textsc{Small} only}:\\
Loss of smell and taste validated as COVID-19 symptoms in patients with high recovery rate.\\
\midrule
\textbf{Claim 2:} COVID-19 hitting some African American communities harder.\\
\midrule
\textbf{Exclude from \textsc{Large} and \textsc{Small}}:\\
The African American community is being hit hard by COVID-19. \\
\textbf{Exclude from \textsc{Small} only}:\\
COVID-19 impacts in African-Americans are different from the rest of the U.S. population.\\
    \bottomrule
    \end{tabular}
    \caption{Claim de-duplication examples.}
    \label{tab:claimDeduplicationEx}
\end{table}

To illustrate the difference between the two versions of the dataset, we present some examples of claims in Table \ref{tab:claimDeduplicationEx}. For Claim 1, the semantically similar claim `\emph{Loss of smell may suggest milder COVID-19}' is identified and  excluded from both \textsc{Large} and \textsc{Small} datasets. But the claim `\emph{Loss of smell and taste validated as COVID-19 symptoms in patients with high recovery rate}', which includes mentions of another symptom and the recovery rate, is only excluded from the \textsc{Small} dataset. For Claim 2, the rephrased claim `\emph{The African American community is being hit hard by COVID-19}' is excluded from both datasets. But the claim `\emph{COVID-19 impacts in African-Americans are different from the rest of the U.S. population}', which refers specifically to the U.S. population, is only excluded from the \textsc{Small} dataset.

\subsection{Dataset Statistics}\label{sec:dataset_desc}

Our final dataset statistics are shown in the lower part of Table \ref{tab:P-dataset}, where the original and the two reduced versions are presented. After the steps described in Section \ref{sec:preprocessing} the \textsc{Large} dataset contains 5,143 claims, and the \textsc{Small} version 1,709 claims. 

\begin{table}[!ht]
    \centering
     \resizebox{\columnwidth}{!}{%
    \begin{tabular}{lrrr}
    \toprule \textbf{Category} & \textbf{Orig.} & \textbf{\textsc{Large}} & \textbf{\textsc{Small}}\\
    \midrule
    Similarity & $0.67\pm 0.23$ & $0.43\pm 0.13$ & $0.37\pm 0.14$ \\
    $\eta_{.90}$ & 0.99 & 0.60 & 0.56 \\
    \midrule
    False & 14,739 & 1,810 & 477 \\
    True & 5,950 & 3,333 & 1,232 \\
    \midrule
    Total & 20,689 & 5,143 & 1,709\\
    \bottomrule
    \end{tabular}}
    \caption{The average claim similarity values and the PANACEA \textsc{Large} and \textsc{Small} dataset statistics. $\eta_{.90}$ denotes the 90th percentile value.}
    \label{tab:P-dataset}
\end{table}

Example claims contained in the dataset are shown in Table \ref{tab:examples-dataset}. Each of the entries in the dataset contains the following information:

\begin{itemize}
\item \emph{Claim}. Text of the claim.
\item \emph{Claim label}. The labels are: \emph{False}, and \emph{True}.
\item \emph{Claim source}. The sources include mostly fact-checking websites, health information websites, health clinics, public institutions sites, and peer-reviewed scientific journals.
\item \emph{Original information source}. Information about which general information source was used to obtain the claim.
\item \emph{Claim type}. The different types, explained in Section \ref{sec:categorisation}, are: \emph{Multimodal, Social Media, Questions, Numerical}, and \emph{Named Entities}.
\end{itemize}

\begin{table*}[ht]\small
    \centering
    \begin{tabular}{p{5cm}rp{3cm}p{2cm}p{2cm}}
    \toprule \textbf{Claim} & \textbf{Category} & \textbf{Source} & \textbf{Orig. data src.} & \textbf{Type}\\
    \midrule
Stroke Scans Could Reveal COVID-19 Infection. & True & ScienceDaily & CoAID & \\
Whiskey and honey cure coronavirus. & False & Independent news site & CovidLies & \\
COVID-19 is more deadly than Ebola or HIV. & False & Australian Associated Press &  Poynter & \\
Dextromethorphan worsens COVID-19. & True & Nature & TREC Health Misinformation track & \\
ACE inhibitors increase risk for coronavirus. & False & Infectious Disorders - Drug Targets journal & TREC COVID challenge & \\
Nancy Pelosi visited Wuhan, China, in November 2019, just a month before the COVID-19 outbreak there. & False & Snopes & MM-COVID & Named Entity, Numerical content\\
    \bottomrule
    \end{tabular}
    \caption{Example entries in the constructed PANACEA dataset.}
    \label{tab:examples-dataset}
\end{table*}

\section{Claim Veracity Assessment}
We develop a pipeline approach consisting of three steps: document retrieval, sentence retrieval and veracity assessment for claim veracity evaluation. Given a claim, we first retrieve the most relevant documents from COVID-19 related sources and then further retrieve the top $N$ most relevant sentences. Considering each retrieved sentence as evidence, we train a veracity assessment model to assign a \emph{True} or \emph{False} label to the claim.

\subsection{Document Retrieval}

\paragraph{Document Dataset.}
\label{sec:inforSourceDataset}

In order to retrieve documents relevant to the claims, we first construct an additional dataset containing documents obtained from reliable COVID-19 related websites. These information sources represent a real-world comprehensive database about COVID-19 that can be used as a primary source of information on the pandemic. We have selected four organisations from which to collect the information: 
    (1) \emph{Centers for Disease Control and Prevention (CDC)}, national public health agency of the United States; (2) \emph{European Centre for Disease Prevention and Control (ECDC)}, EU agency aimed at strengthening Europe's defenses against infectious diseases; (3) \emph{WebMD}, online publisher of news and information on health; and (4) \emph{World Health Organization (WHO)}, agency of the United Nations responsible for international public health.

All pages corresponding to the COVID-19 sub-domains of each site have been downloaded. The web content was downloaded using the BeautifulSoup\footnote{\url{https://www.crummy.com/software/BeautifulSoup/}} and Scrapy\footnote{\url{https://scrapy.org/}} packages. Social networking sites and non-textual content were discarded. In total 19,954 web pages have been collected. The list of websites and the full content of each website constitute this additional dataset used for document retrieval. 
This dataset is enhanced with some additional websites used only in the document retrieval experiments, detailed in Section \ref{sub:info_retr}.

\paragraph{Method.}
\label{sec:docretrieval}
Information sources were indexed by creating a Pyserini Lucene index and PyGaggle was used to implement a re-ranker model on the results. The documents were split into paragraphs of 300 tokens segmented with a BERT tokenizer.

To retrieve the information we first used a BM25 score. Additionally, we tested the effect of multi-stage retrieval by re-ranking the initial results using MonoBERT \citep{nogueira2019multi} and MonoT5 models, and query expansion using RM3 pseudo-relevance feedback \citep{abdul2004umass} on the BM25 results \citep{lin2019neural,yang2019critically}.

MonoBERT uses a BERT model trained using as inputs the query and each of the documents to be re-ranked encoded together (\texttt{[CLS]query[SEP]doc[SEP]}), and then the \texttt{[CLS]} output token is passed to a single layer fully-connected network that produces the probability of the document being relevant to the query.

\subsection{Sentence Retrieval}
\label{sec:sentenceRetrieval}

For each claim, once documents are retrieved using BM25 and MonoT5 re-ranking of the top 100 BM25 results, we then further retrieve the $N$ most similar sentences obtained from the 10 most relevant documents. 
The relevance of the sentences is calculated using cosine similarity in relation to the original claim. The similarity is obtained with the pre-trained model MiniLM-L12-v2 \citep{wang2020minilm}, using Sentence-Transformers\footnote{\label{ukplab}\url{https://github.com/UKPLab/sentence-transformers}} \citep{reimers-2019-sentence-bert} to encode the sentences.

\subsection{Veracity Assessment}
\label{sec:veracity}

\begin{figure*}[ht]
     \centering
     \begin{subfigure}[b]{\textwidth}
         \centering
        \includegraphics[width=0.95\linewidth]{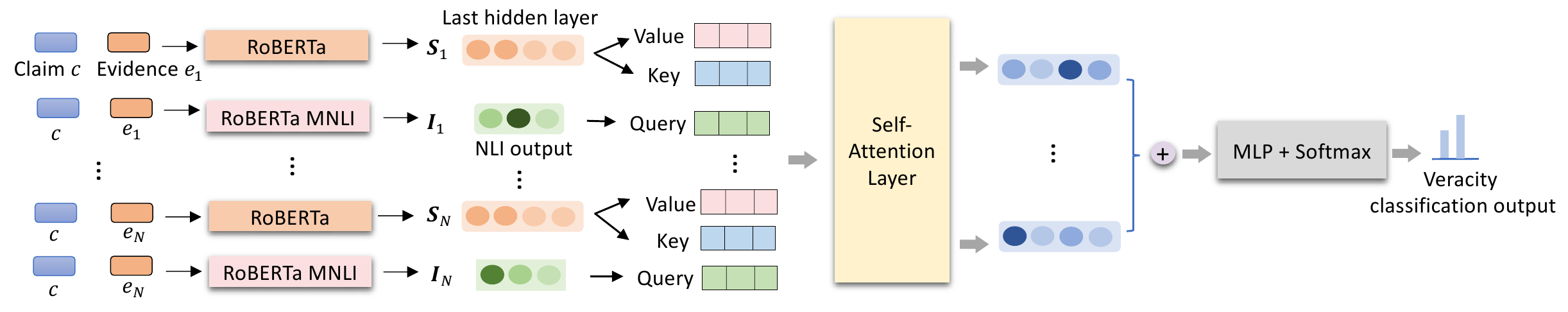}
         \caption{\texttt{NLI-SAN}}
         \label{fig:NLI-SAN}
     \end{subfigure}
    \vfill
     \begin{subfigure}[b]{\textwidth}
         \centering
        \includegraphics[width=0.95\linewidth]{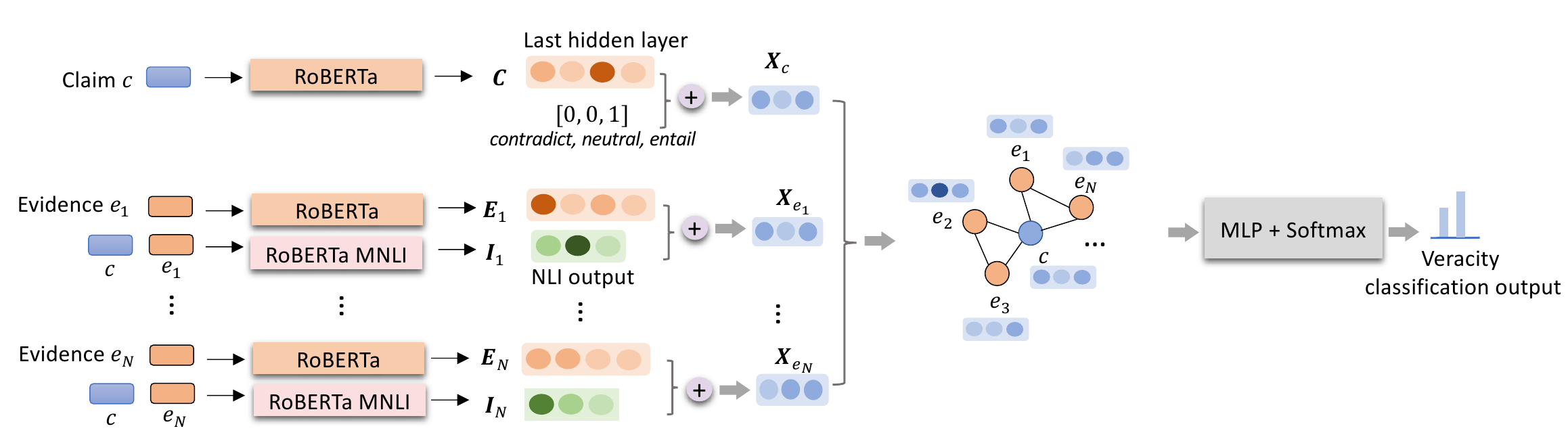}
         \caption{\texttt{NLI-graph}}
         \label{fig:NLI-graph}
     \end{subfigure}
    \caption{Proposed veracity classification models. $\oplus$ means concatenation.}
    \label{fig:veracityModels}
\end{figure*}

We propose two veracity assessment approaches built on the NLI results of claim-evidence pairs. For each of the most similar sentences (pieces of evidence) retrieved for a claim, we apply the pre-trained NLI model RoBERTa-large-MNLI\footnote{\label{huggface}\url{https://huggingface.co/}}\citep{liu2019roberta}. This model acts as a cross-encoder on pairs of sentences, trained to detect the relationship between the two sentences: \emph{contradiction}, \emph{neutrality}, or \emph{entailment}. The model is trained on the Multi-Genre Natural Language Inference (MultiNLI) dataset \citep{N18-1101}. The inference results are then used in our proposed approaches described below.

\paragraph{NLI-SAN.}The first approach, named \texttt{NLI-SAN}, incorporates the inference results of claim-evidence pairs into a Self-Attention Network (SAN) (See Figure \ref{fig:NLI-SAN}). First, a claim is paired with each piece of retrieved relevant evidence. Each pair $(c,e_i)$ is fed into a RoBERTa-large\textsuperscript{\ref{huggface}} model, and the last hidden layer output $\mathbf{S_i}$ is used as its representation. Additionally, each pair is also fed to the mentioned RoBERTa-large-MNLI\textsuperscript{\ref{huggface}} model obtaining $\mathbf{I_i}$, a triplet containing the probability of \emph{contradiction}, \emph{neutrality}, or \emph{entailment}. 
\begin{equation}
\begin{split}
\mathbf{S_i} &= \text{RoBERTa}(c,e_i) \\
\mathbf{I_i} &= \text{RoBERTa}_{\text{NLI}}(c,e_i)
\end{split}
\end{equation}

\noindent The sentence representation is combined with the NLI output through a Self Attention Network (SAN) \citep{galassi2020attention,bahdanau2014neural}. 

The RoBERTa-encoded claim-evidence representation $\mathbf{S_i}$ with length $n_S = n_K = n_V$ is mapped onto a Key $\mathbf{K}\in\mathbb{R}^{n_K\times d_K}$ and a Value $\mathbf{V}\in\mathbb{R}^{n_V\times d_V}$, while the NLI output $\mathbf{I_i}$ of each claim-evidence pair is mapped onto a Query $\mathbf{Q}\in\mathbb{R}^{n_Q\times d_Q}$. The representation dimensionality is $d_K = d_V = d_Q = 1024$. The attention function is defined as:
\begin{equation}
\text{Att}(\mathbf{Q},\mathbf{K},\mathbf{V}) = \text{softmax}(\mathbf{Q} \mathbf{K^\top}/\sqrt d) \mathbf{V}
\end{equation}

\noindent While standard attention mechanisms use only the sentence representation information for the Key, Value and Query, here the inference information is used in the Query. This \emph{attention} mechanism is applied to each of the claim-evidence pairs, and the outputs are concatenated into an output $\text{O}_{\text{SAN}}$ that is passed through a Multi-Layer Perceptron (MLP) with hidden size $d_h$ and a Softmax layer to generate the veracity classification output.
\begin{equation}
\hat{\bm{y}} = \text{softmax}(\text{MLP}_{\text{ReLU}}(\text{O}_{\text{SAN}}))
\end{equation}

\paragraph{NLI-graph.}
We propose an alternative approach based on Graph Convolutional Networks (GCN). First, for each claim-evidence pair, we derive RoBERTa-encoded representations for the claims and evidence separately (using the pooled output of the last layer) and obtain NLI results of the pairs as before. 
\begin{gather}
\mathbf{C_i} = \text{RoBERTa}(c);\quad \mathbf{E_i} = \text{RoBERTa}(e_i) \\
\mathbf{I_i} = \text{RoBERTa}_{\text{NLI}}(c,e_i)
\end{gather}

\noindent Next, we build an evidence network in which the central node is the claim and the rest of the nodes are the evidence. Two nodes are linked if their similarity value exceeds a pre-defined threshold, which is empirically set to 0.9 by comparing the results of the experimental evaluation described in the following section using different thresholds. The similarity is considered between claim and evidence, but also between pieces of evidence. Similarity calculation is performed following the same approach as in Section \ref{sec:sentenceRetrieval}. The features considered in each evidence node are the concatenation of $\mathbf{E_i}$ and $\mathbf{I_i}$. For the claim node we use its representation $\mathbf{C_i}$ and a unity vector $(0,0,1)$ for the inference. The network is implemented with the package PyTorch Geometric \citep{Fey/Lenssen/2019}, using in the first layer the GCNConv operator \citep{kipf2016semi} with 50 output channels and self-loops to the nodes, represented by: 
\begin{equation}
\mathbf{X}^{\prime} = \mathbf{\hat{D}}^{-1/2} \mathbf{\hat{A}}
\mathbf{\hat{D}}^{-1/2} \mathbf{X} \mathbf{W},
\end{equation}
where $\mathbf{X}$ is the matrix of node feature vectors, $\mathbf{\hat{A}} = \mathbf{A} + \mathbf{I}$ denotes the adjacency matrix with inserted self-loops, $\hat{D}_{ii} = \sum_{j=0} \hat{A}_{ij}$ its diagonal degree matrix, and $\mathbf{W}$ is a trainable weight matrix.

Once the node representation is updated via GCN, all the node representations are averaged and passed to the MLP and the Softmax layer to generate the final veracity classification output.
\begin{equation}
\hat{\bm{y}} = \text{softmax}(\text{MLP}_{\text{ReLU}}(\text{O}_{\text{graph}}))
\end{equation}

\section{Experiments}

In this section, we perform a twofold evaluation:
We first evaluate our document retrieval methods (presented in \textsection{\ref{sec:docretrieval}}) on obtaining information relevant to the dataset claims from a database of COVID-19 related websites. We subsequently present an evaluation of the veracity assessment approaches for the claims (described in \textsection{\ref{sec:veracity}}).

\subsection{Document Retrieval}\label{sub:info_retr}

In order to evaluate our document retrieval methods, we need the gold-standard relevant document for each claim. Therefore, in the documents dataset described in section \ref{sec:inforSourceDataset} we additionally include the web content referenced in each of the information sources used to compile our claim dataset:

\noindent\underline{The CoronaVirus Alliance Database}. All web pages from the websites referenced as fact-checking sources for the claims have been downloaded from 151 different domains.

\noindent\underline{CoAID dataset}. We downloaded the websites used as fact-checking sources of false claims and the websites where correct information on true claims is gathered from 68 different domains.

\noindent\underline{MM-COVID}. We collected both fact-checking sources and reliable information related to the claims of this dataset from 58 web domains.

\noindent\underline{CovidLies dataset}. We include the web content used as fact-checking sources of the misconceptions from 39 domains.

We have not included web content from the TREC Challenges, as each of them is performed on a very large dataset specific to each challenge (CORD19 and Common Crawl corpus), as explained previously.
Note that in our subsequent experiments, we have excluded all fact-checking websites to avoid finding directly the claim references.
The results of the document retrieval are presented in Table \ref{tab:IR-results}. For each claim, the precision@$k$ is defined as 1 if the relevant result is retrieved in the top $k$ list and 0 otherwise.

\begin{table}[!ht]
    \centering
    \resizebox{\columnwidth}{!}{%
    \begin{tabular}{lrrrr}
    \toprule  & \textbf{AP@5} & \textbf{AP@10} & \textbf{AP@20} & \textbf{AP@100}\\
    \midrule
    BM25 & 0.54 & 0.56 & 0.58 & 0.62 \\
    BM25+MonoBERT & 0.52 & 0.55 & 0.58 &  0.62 \\
    BM25+MonoT5 & \textbf{0.55} & \textbf{0.58} & \textbf{0.60} & \textbf{0.62} \\
    BM25+RM3+MonoT5 & 0.51 & 0.53 & 0.55 & 0.57 \\
    \bottomrule
    \end{tabular}}
    \caption{Document retrieval results. Average precision for different cut-offs. For the MonoBERT and MonoT5 cases, 100 initial results are retrieved in the first retrieval stage before re-ranking.}
    \label{tab:IR-results}
\end{table}

We can see that by using BM25, it is possible in many cases to retrieve the relevant results at the very top of our searches. Combining BM25 with MonoBERT did not offer any improvement. It even introduced noise to the retrieval results, leading to inferior performance compared to using BM25 only on AP@5 and AP@10. MonoT5 appears to be more effective, consistently improving the retrieval results across all metrics. 
Moreover for this dataset the use of query expansion using RM3 pseudo-relevance feedback on the BM25 results does not improve the results.
 
\subsection{Veracity Assessment Evaluation}\label{sub:ver}

\begin{table*}[!ht]
    \centering\small
    \begin{tabular}{lrrrrrrr}
    \toprule
    \multirow{2}{*}{\textbf{Model}} & \multicolumn{3}{c}{\textbf{False}} & \multicolumn{3}{c}{\textbf{True}} & \multirow{2}{*}{\textbf{Macro F1}}\\
   \cmidrule(lr){2-4} \cmidrule(lr){5-7} 
     & \textbf{Precision} & \textbf{Recall} & \textbf{F1} & \textbf{Precision} & \textbf{Recall} & \textbf{F1} & \\
    \midrule
    GEAR \citep{zhou2019gear} & 0.81 & 0.60 & 0.69 & 0.85 & 0.94 & 0.89 & 0.79 \\
    KGAT \citep{liu2019fine} & 0.89 & \textbf{0.96} & \textbf{0.92} & \textbf{0.98} & 0.95 & \textbf{0.97} & \textbf{0.94} \\
    \midrule
    \texttt{NLI}  & 0.48 & 0.24 & 0.31 & 0.75 & 0.90 & 0.82 & 0.56\\
    \texttt{NLI+Sent}  & 0.91 & 0.87 & 0.89 & 0.95 & \textbf{0.97} & 0.96 & 0.92\\
    \texttt{NLI+PSent}  & 0.87 & 0.72 & 0.79  & 0.90 & 0.96 & 0.93 & 0.86\\
    \texttt{NLI-SAN}  & \textbf{0.93} & 0.89 & 0.91  & 0.96 & \textbf{0.97} & \textbf{0.97} & \textbf{0.94}\\
    \midrule
    \texttt{NLI-graph$_{-abl}$}  & 0.50 & 0.33 & 0.39 & 0.77 & 0.87 & 0.81 & 0.60\\
    \texttt{NLI-graph} & 0.89 & 0.83 & 0.86  & 0.94 & 0.96 & 0.95 & 0.90\\
    \bottomrule
    \end{tabular}
    \caption{Veracity classification results on the PANACEA \textsc{Small} dataset. The best result in each column is highlighted in bold.}
    \label{tab:multi_results}
\end{table*}

\begin{table*}[!ht]
    \centering\small
    \begin{tabular}{lrrrrrrr}
    \toprule
    \multirow{2}{*}{\textbf{Model}} & \multicolumn{3}{c}{\textbf{False}} & \multicolumn{3}{c}{\textbf{True}} & \multirow{2}{*}{\textbf{Macro F1}}\\
   \cmidrule(lr){2-4} \cmidrule(lr){5-7} 
     & \textbf{Precision} & \textbf{Recall} & \textbf{F1} & \textbf{Precision} & \textbf{Recall} & \textbf{F1} & \\
    \midrule
    GEAR \citep{zhou2019gear} & 0.88 & 0.88 & 0.88 & 0.93 & 0.94 & 0.94 & 0.91 \\
    KGAT \citep{liu2019fine} & \textbf{0.95} & \textbf{0.98} & \textbf{0.96} & \textbf{0.99} & \textbf{0.98} & \textbf{0.98} & \textbf{0.97} \\
    \midrule
    \texttt{NLI} & 0.52 & 0.27 & 0.36 & 0.69 & 0.86 & 0.76 & 0.56 \\
    \texttt{NLI+Sent} & 0.94 & 0.94 & 0.94 & 0.97 & 0.97 & 0.97 & 0.95 \\ 
    \texttt{NLI+PSent} & 0.89 & 0.77 & 0.82 & 0.88 & 0.95 & 0.91 & 0.86 \\
    \texttt{NLI-SAN} & \textbf{0.95} & 0.95 & 0.95 & 0.97 & \textbf{0.98} & 0.97 & 0.96 \\
    \midrule
    \texttt{NLI-graph$_{-abl}$} & 0.60 & 0.43 & 0.50 & 0.73 & 0.84 & 0.78 & 0.64 \\
    \texttt{NLI-graph} & 0.94 & 0.91 & 0.93 & 0.95 & 0.97 & 0.96 & 0.94 \\
    \bottomrule
    \end{tabular}
    \caption{Veracity classification results on the PANACEA \textsc{Large} dataset. The best result in each column is highlighted in bold.}
    \label{tab:multi_results_large}
\end{table*}

Here we evaluate our proposed \texttt{NLI-SAN} and \texttt{NLI-graph} veracity assessment approaches. To gain a better insight into the benefits of the proposed architectures, we conducted additional experiments on the variants of the models including:

\begin{itemize}[label=\textbullet, nolistsep, noitemsep]
\item \texttt{NLI}, using only the NLI outputs of the claim-evidence pairs. The outputs are concatenated and then passed through the final classification layer to generate veracity classification results.

\item \texttt{NLI+sent}, this is the ablated version of \texttt{NLI-SAN} without the self-attention layer. Here, the RoBERTa-encoded claim-evidence representations are concatenated with the NLI results and then fed to the classification layer to produce the veracity classification output. 

\item \texttt{NLI+PSent}, this is similar to the previous ablated version, but using the pooled representation of the claim-evidence pair to concatenate with the NLI result.

\item \texttt{NLI-graph$_{-abl}$}, this is the ablated version of \texttt{NLI-graph} in which the node representation is the NLI result of the corresponding claim-evidence pair without its RoBERTa-encoded representation.
\end{itemize}

For \texttt{NLI}, \texttt{NLI+sent} and \texttt{NLI-SAN}, we consider the 5 most similar sentences for each claim, obtained from the 10 most relevant documents of the information source database. Those documents are retrieved using BM25 and MonoT5 re-ranking of the top 100 BM25 results. For \texttt{NLI-graph}, \texttt{NLI-graph$_{-abl}$} and \texttt{NLI+PSent}, in order to have enough nodes to benefit from the network structure, the number of retrieved sentences is increased to 30 for each claim, selected as the 3 most similar sentences from the top 10 retrieved documents. The retrieval procedure is as in sections~\ref{sec:docretrieval} and ~\ref{sec:sentenceRetrieval}. Details of parameter settings can be found in Appendix \ref{sec:parameter_setting}. 
We compare against the SOTA methods GEAR\footnote{\url{https://github.com/thunlp/GEAR}}\citep{zhou2019gear} and KGAT\footnote{\url{https://github.com/thunlp/KernelGAT}}\citep{liu2019fine}, with settings as described by the authors.

For all approaches we perform 5-fold cross-validation and report the averaged results on the \textsc{Small} dataset in Table \ref{tab:multi_results}.
By using the NLI information alone it is possible to obtain reasonable results for the \emph{True} claims, however, this is not the case for the most relevant \emph{False} claims. Once we add sentence representations the efficiency of the method increases significantly. Using \texttt{NLI-SAN} instead of simply concatenating contextualised claim-evidence representations and NLI outputs further improves the results. A similar observation can be made in the results generated by \texttt{NLI-graph} and its variants; the contextualised representations of claim-evidence pairs are much more important than merely using the corresponding NLI values. We also note that using the graph version \texttt{NLI-graph} obtains better scores than a non-graph model with the same information \texttt{NLI+PSent}, however the scores are still lower than the \texttt{NLI-SAN} method. Our method performs on a par with KGAT, while being simpler, and outperforms GEAR.

Complementing the results for the \textsc{Small} dataset, Table \ref{tab:multi_results_large} presents the results for the \textsc{Large} dataset. In general, we observe improved performance for all models across all metrics for both classes compared to the results on the \textsc{Small} dataset. The previous results in the \textsc{Small} dataset constitute a more challenging case, since the uniqueness of the claims is increased and therefore the veracity assessment models are not able to learn from similar claims when performing the assessment.
 
\subsection{Discussion}
\label{discussion}
Our results show that in document retrieval, we have obtained values of around 0.6 from a simple term scoring and re-ranking retrieval model. However, this baseline represents only a rough measure of quality using this technique, since we have only evaluated the retrieval of a single document specific to each claim; we have not evaluated the quality of other retrieved documents.

The distinction into \emph{True} and \emph{False} claims can be rather coarse-grained. We note that initially we considered a larger number of veracity labels, including more nuanced cases that could be interesting to analyse (see \ref{appendix_preproc}). However, we have not found a clear separation between complex cases and it would seem that different fact checkers do not follow the same conventions when labelling such cases. The development of datasets especially focused on such nuanced cases may be therefore an important line of work in the future, together with the development of techniques for these more complex situations. 

In analysing misclassified claims, we note some interesting cases. The scope and globality of the pandemic imply that similar issues are mentioned repeatedly on multiple occasions, yet claims to be verified may include nuances or specificities. This is challenging as it is easy to retrieve information that omits relevant nuances. E.g. The claim ``\emph{Barron Trump had COVID-19, Melania Trump says}" retrieves sentences such as ``\emph{Rudy Giuliani has tested positive for COVID-19, Trump says.}" with a similar structure and mentions but missing the key name. This type of situation could be addressed by using Named Entity Recognition (NER) methods that prioritise matching between the entities involved in the claim and the information sources. See e.g. \citep{taniguchi2018integrating, nooralahzadeh2018sirius}.

Other interesting cases involve claims for which documents with adequate information are retrieved, but the sentences containing evidence cannot be identified because they are too different from the original claim. E.g. The claim ``\emph{Vice President of Bharat Biotech got a shot of the indigenous COVAXIN vaccine}" retrieves correct documents on the issue. Similar sentences are retrieved such as ``\emph{Covaxin which is being developed by Bharat Biotech is the only indigenous vaccine that is approved for emergency use.}". Despite being similar such retrieved sentences give no information about the claimed situation. In the retrieved document, the sentence ``\emph{The pharmaceutical company, has in a statement, denied the claim and said the image shows a routine blood test.}" contains the essential information to debunk the original claim but is missed by the sentence retrieval engine as it is very different from the claim (See Table \ref{tab:misid-dataset} in Appendix \ref{sec:additional_errors} for other examples).

Such cases are more difficult to deal with, as the similarity between claim and evidence is certainly a good indicator of relevance. Nevertheless, these cases are very interesting for future work using more complex approaches. We have made an initial attempt to address this problem by representing claims and retrieved documents using Abstract Meaning Representation \citep{banarescu2013abstract} in order to better select relevant information.
Although the results were not satisfactory, it may be an interesting avenue for future exploration. Another line of future work is the design of strategies against adversarial attacks to mitigate possible risks to our system.

\section{Conclusions}
We have presented a novel dataset that aggregates a heterogeneous set of COVID-19 claims categorised as \emph{True} or \emph{False}. Aggregation of heterogeneous sources involved a careful deduplication process to ensure dataset quality.
Fact-checking sources are provided for veracity assessment, as well as additional information sources for \emph{True} claims. Additionally, claims are labelled with sub-types (Multimodal, Social Media, Questions, Numerical, and Named Entities).

We have performed a series of experiments using our dataset for information retrieval through direct retrieval and using a multi-stage re-ranker approach. We have proposed new NLI methods for claim veracity assessment, attention-based \texttt{NLI-SAN} and graph-based \texttt{NLI-graph}, achieving in our dataset competitive results with the GEAR and KGAT state-of-the-art models. We have also discussed challenging cases and provided ideas for future research directions.

\section*{Acknowledgements}
This work was supported by the UK Engineering and Physical Sciences Research Council (grant no. EP/V048597/1, EP/T017112/1). ML and YH are supported by Turing AI Fellowships funded by the UK Research and Innovation (grant no. EP/V030302/1, EP/V020579/1).

\bibliographystyle{acl_natbib}
\bibliography{factverification}

\clearpage
\appendix
\setcounter{table}{0}
\renewcommand{\thetable}{A\arabic{table}}

\section{Data Sources}
\label{appendix_sources}

Here we present detailed information of the data sources introduced in section \ref{sec:dataSource}.

It is worth noting that for the construction of our dataset, we have only included sources or datasets that contain explicit veracity labels of specific claims, thus we have not included collections of tweets related to COVID that do not have veracity labels \citep{chen2020tracking, lamsal2021design, abdul2020mega, huang_xiaolei_2020_3735015, dimitrov2020tweetscov19, kerchner2020coronavirus, qazi2020geocov19}. We have not included claims without independent fact-checking sources \citep{memon2020characterizing,shahi2021exploratory} and information sources without formulated claims such as the collections of scholarly articles \citep{wang2020cord, chen2021litcovid}, news articles \citep{zhou2020recovery}, or articles obtained through specific repositories as \citep{portaleu, portalwho, portalelsevier, portalcambridge, portalancet, portaloxford, portalmedrn}.

The data sources that we have used for the construction of our dataset are:

\begin{itemize}
    \item \textbf{The CoronaVirusFacts/DatosCoronaVirus Alliance Database\footnote{\url{https://www.poynter.org/ifcn-covid-19-misinformation/}}}. Published by Poynter\footnote{\url{www.poynter.org}}, this online publication combines fact-checking articles from more than 100 fact-checkers from all over the world, being the largest journalist fact-checking collaboration on the topic worldwide\footnote{\url{https://www.poynter.org/coronavirusfactsalliance/}}. The publication is presented as an online portal, thus we had to develop scripts to crawl the content and extract the relevant claims, categories, and information sources.

    \item \textbf{CoAID dataset\footnote{\url{https://github.com/cuilimeng/CoAID}}}. The dataset \citep{cui2020coaid} contains fake news from fact-checking websites and real news from health information websites, health clinics, and public institutions. Unlike most other datasets, it contains a wide selection of true claims.

    \item \textbf{MM-COVID\footnote{\url{https://github.com/bigheiniu/MM-COVID}}}. The multilingual dataset \citep{li2020mmcovid} contains fake and true news collected from Poynter and Snopes\footnote{\url{www.snopes.com}}, being a good complement to the first data source.

    \item \textbf{CovidLies dataset\footnote{\url{https://github.com/ucinlp/covid19-data}}}. The dataset \citep{hossain2020covidlies} contains a curated list of common misconceptions about COVID appearing in social media, carefully reviewed to contain very relevant and unique claims unlike other automatically collected datasets.

    \item \textbf{TREC Health Misinformation track\footnote{\url{https://trec-health-misinfo.github.io/}}}. Research challenge using claims on the health domain focused on information retrieval from general websites through the Common Crawl corpus\footnote{ \url{https://commoncrawl.org/}}. This dataset is specialized in a very specific domain, and has been used for a very different application than the previous data sources.

    \item \textbf{TREC COVID challenge\footnote{\url{https://ir.nist.gov/covidSubmit/data.html}}}. Research challenge \cite{voorhees2021trec,roberts2020trec} using claims on the health domain focused on information retrieval from scholarly peer-reviewed journals through the CORD19 dataset \citep{wang2020cord}, the largest existing compilation of such articles. Similar to the last source, but focused on scholarly papers unlike the other sources.

\end{itemize}

\subsection{Pre-processing}
\label{appendix_preproc}

A separate pre-processing step was carried out for each of the selected data sources:

\noindent\textbf{The CoronaVirusFacts/CoronaVirus Alliance Database.} The data was downloaded on 13 February 2021. From the 11,647 entries initially obtained, entries with no fact-checking source and categories with less than 10 entries were removed. The different fact-checkers used different categories to label the claims, although in most of the cases the difference was mainly in terms of spelling. Initially we identified the following common categories:
\emph{False} (including FALSE, FALSO, Fake, false, false and misleading, Two Pinocchios, Misinformation / Conspiracy theory, Not true, false headline, MANIPULATED, Unproven), \emph{Misleading} (MIsleading, MISLEADING, mislEADING, MiSLEADING, misleading, Misleading/False, Misleading), \emph{Missing Context} (Missing context, Needs Context, missing context), \emph{No Evidence} (NO EVIDENCE, No evidence, No Evidence), \emph{Mostly False} (Mostly False, Mostly false, MOSTLY FALSE, mostly false, Mainly false), \emph{Partially False} (Partially False, Partly false, Partially false, partially false, partly false), \emph{Partially True} (PARTLY TRUE, Partially correct, Partially true, Partly true, HALF TRUE, HALF TRUTH, half true), and \emph{Mostly True} (Mostly true, MOSTLY TRUE, mainly correct).
Next, we conducted a manual inspection of the different categories. We found that the categories \emph{Misleading, Missing Context, No Evidence, Mostly False}, and \emph{Partially False} had no homogeneous and clear definition through the different fact-checking media. Each group contains claims fitting the definition mixed with claims that are simply false (e.g. of false claims under other labels: ``Misleading: Only people from South Korea have Covid-19 antibodies"; ``Partially False: The vaccines contain substances such as arsenic or uranium according to scientific studies"; ``Mostly False: Pope Francis contracted coronavirus"). Therefore, we decided to not use these nuanced categories but group them in the general \emph{False} category. Additionally, we removed the 25 claims from the categories \emph{Partially True} and \emph{Mostly True}, since they contained both \emph{True} and \emph{False} claims.

\noindent\textbf{CoAID dataset.} The datasets \texttt{NewsRealCOVID-19}, \texttt{NewsFakeCOVID-19}, and \texttt{ClaimFakeCOVID-19} were selected. The additional available dataset contains claims already existing in other datasets, formulated in this case as questions, and thus was not included. The selected datasets contain \emph{True} and \emph{False} claims.

\noindent\textbf{MM-COVID.} The claims were obtained from the \texttt{English\_news} part of the dataset since we are only interested in English claims. 3,409 claims were collected. Claims in other languages appeared in the file, therefore we did a language filtering using polyglot\footnote{\url{https://github.com/aboSamoor/polyglot}}. Additionally, claims without fact-checking sources were deleted. It contains \emph{True} and \emph{False} claims.

\noindent\textbf{CovidLies dataset.} The available claims have been manually revised by eliminating duplicates, resulting in a total of 62 misconception claims. It contains \emph{False} claims.

\noindent\textbf{TREC Health Misinformation track.} The claims used in the track were obtained and reformulated manually by us as affirmative claims (e.g., ``\emph{Can vitamin D cure COVID-19?}" was changed to ``\emph{Vitamin D cures COVID-19}") for consistency with the rest of the data sources and to allow claim veracity assessment. \emph{True} and \emph{False} claims are used.

\noindent\textbf{TREC COVID challenge.} The claims used in the challenge were obtained and reformulated manually by us as full sentences using the explanations related to each query (e.g., for a given query ``\emph{coronavirus immunity}", and its explanation ``\emph{will SARS-CoV2 infected people develop immunity?}", we form the following claim, ``\emph{coronavirus infected people develop immunity}"). \emph{True} and \emph{False} claims are used.

The above processed data sources were combined to provide 20,689 initial claims.

\subsection{Claim Categorisation}
\label{sec:categorisation}

The claims were analysed to identify types of claims that may be of particular interest, either for inclusion or exclusion depending on the type of analysis. The following types were identified: (1) Multimodal; (2) Social media references; (3) Claims including questions; (4) Claims including numerical content; (5) Named entities, including: PERSON $-$ People, including fictional; ORGANIZATION $-$ Companies, agencies, institutions, etc.; GPE $-$ Countries, cities, states; FACILITY $-$ Buildings, highways, etc. These entities have been detected using a RoBERTa base English model \citep{liu2019roberta} trained on the OntoNotes Release 5.0 dataset \citep{weischedel2013ontonotes} using Spacy\footnote{ \url{https://spacy.io/}}.

\begin{table*}[ht!]\small
    \centering
    \begin{tabular}{p{4cm}p{10cm}}
    \toprule \textbf{Claim} & \textbf{Description}\\
    \midrule
 ``\emph{Sugar causes a cytokine storm in the lungs that promotes COVID-19}"  & Retrieved documents are relating COVID and its cytokine storm effects, but without the specific mention of sugar, which does not cause a cytokine storm.\\
 \midrule
``\emph{Barron Trump had COVID-19, Melania Trump says}" & Retrieved sentences such as ``\emph{Rudy Giuliani has tested positive for COVID-19, Trump says.}" with a similar structure and mentions but mistaking the family members and missing the key name. \\
\midrule
``\emph{Prince Charles tested positive for COVID-19 after meeting Bollywood singer Kanika Kapoor.}" & Documents mentioning Prince Charles positive COVID tests are obtained, but without any mentions to the singer.\\
\midrule
``\emph{Vice President of Bharat Biotech got a shot of the indigenous COVAXIN vaccine}" & Correct documents on the issue are retrieved. Similar sentences are retrieved such as ``\emph{Covaxin which is being developed by Bharat Biotech is the only indigenous vaccine that is approved for emergency use.}" or ``\emph{Bharat Biotech's Covaxin is the first Indian vaccine to receive approval to conduct Phase I/Phase II trials.}". However, being similar they give no information about the claimed situation. In the retrieved document, the sentence ``\emph{The pharmaceutical company, has in a statement, denied the claim and said the image shows a routine blood test.}" contains the essential information to debunk the original claim. But it is missed by the sentence retrieval engine as it is very different from the claim. \\
\midrule
``\emph{Masks can be sanitized in microwave}" & Correct documents are retrieved with similar sentences such as ``\emph{Claiming masks can be sanitized in microwave resurfaces}". However, sentences such as ``\emph{The study authors cautioned health care workers against trying to clean masks this way. Microwaves melted the masks, making them useless.}" or ``\emph{He also warns people against using microwaves or ovens to heat their masks.}" that are present in the retrieved documents but are not similar enough to the claim are missed. \\

    \bottomrule
    \end{tabular}
    \caption{Examples of errors in document or sentence retrieval.}
    \label{tab:misid-dataset}
\end{table*}

\section{Parameter Setting}
\label{sec:parameter_setting}

In our veracity assessment experiments, the parameters of the initial RoBERTa models are frozen during the training. 
The inputs are padded and truncated to the longest sequence, and a ReLU function is used as the activation function for the hidden layer. The GCNConv outputs are padded to the longest graph size. The loss function used is cross-entropy. The size of the hidden layer is 50, the batch size is 30, and the training is performed for 100 epochs for \texttt{NLI-SAN} and its variants, and 200 epochs for \texttt{NLI-graph} and its variants. The optimizer used is AdamW \citep{loshchilov2017decoupled} with $\beta_1=0.9$, $\beta_2=0.999$, a weight decay of 0.01, and a learning rate of $10^{-2}$ for \texttt{NLI}, $10^{-4}$ for \texttt{NLI+Sent} and \texttt{NLI-SAN}, and   a learning rates of $10^{-4}$ for \texttt{NLI-graph}, $10^{-3}$ for \texttt{NLI-graph$_{-abl}$}, and $10^{-5}$ for \texttt{NLI+PSent}, these last three with a step size of $0.1$ after 100 epochs.

\section{Additional Examples of Document or Sentence Retrieval Errors}\label{sec:additional_errors}

Here we expand on the examples mentioned in Section \ref{discussion} related to difficulties in the document or sentence retrieval parts of the process. Table \ref{tab:misid-dataset}  presents in more detail the cases previously mentioned, and includes new examples.

\end{document}